%% file: main.tex
\documentclass{article}
\usepackage{spconf,amsmath,graphicx}
\usepackage{colortbl}
\usepackage{booktabs}
\usepackage{amssymb}
\usepackage{bbding}
\usepackage{diagbox}
\usepackage{bbm}
\usepackage{parskip}
\usepackage{makecell}
\usepackage{multirow}
\usepackage{array}
\usepackage{rotating}
\usepackage{hhline}

\usepackage{subcaption}
\usepackage[dvipsnames]{xcolor}
\newcommand{\Xuanlong}[1]{\textcolor{black}{#1}}

\newcommand{\round}[1]{\ensuremath{\lfloor#1\rceil}}
\newcommand{\second}{\cellcolor{blue!10}}
\newcommand{\first}{\cellcolor{blue!30}}
\newcommand{\rotr}[1]{\rotatebox{90}{#1}}

\newcommand\blfootnote[1]{%
\begingroup
\renewcommand\thefootnote{}\footnote{#1}%
\addtocounter{footnote}{-1}%
\endgroup
}

\title{\Xuanlong{On Monocular Depth Estimation and Uncertainty Quantification using Classification Approaches for Regression}}
\name{Xuanlong Yu$^{1,2}$ \qquad Gianni Franchi$^{2}$ \qquad Emanuel Aldea$^{1}$}
\address{$^1$SATIE, Paris-Saclay University \qquad $^2$U2IS, ENSTA Paris, Institut Polytechnique de Paris}

\begin{document}

%\ninept
%
\maketitle
\begin{abstract}
Monocular depth is important in many tasks, such as 3D reconstruction and autonomous driving. Deep learning based models achieve state-of-the-art performance in this field. 
A set of novel approaches for estimating monocular depth consists of transforming the regression task into a classification one. 
\Xuanlong{However, there is a lack of detailed descriptions and comparisons for Classification Approaches for Regression (CAR) in the community and no in-depth exploration of their potential for uncertainty estimation.} 
To this end, this paper will introduce a taxonomy and summary of CAR approaches, a new uncertainty estimation solution for CAR, and a set of experiments on depth accuracy and uncertainty quantification 
\Xuanlong{for CAR-based models on KITTI dataset. The experiments reflect the differences in the portability of various CAR methods on two backbones. Meanwhile, the newly proposed method for uncertainty estimation can outperform the ensembling method with only one forward propagation.}
\end{abstract}
\vspace{-0.5em}
\begin{keywords}
Depth estimation, Uncertainty Estimation
\end{keywords}
\blfootnote{We acknowledge the support of the Saclay-IA computing platform.}
\vspace{-1.5em}
\section{Introduction}
\vspace{-0.5em}
\label{sec:intro}
\input{1introduction}
\vspace{-1.2em}
\section{Overview of Classification Approaches for Regression (CAR)}
\label{sec:survey_method}
\input{3survey_method}
\vspace{-1.5em}
\section{Experiments}
\label{sec:experiments}
\input{4experiments}
\vspace{-1.5em}
\section{Conclusion}\label{sec:Conclusion}
\vspace{-1.0em}
In this paper, we summarize the CAR MDE methods in detail along three key components, and conduct experiments on their portability and performance, including both depth and uncertainty quality. In the future, we will try to apply CAR strategy and E-Dist uncertainty estimator on more tasks.

% -------------------------------------------------------------------------
\clearpage

% \small
\bibliographystyle{IEEEbib}
\bibliography{strings,refs}

\clearpage

\input{5supp}

\end{document}

%% file: 1introduction.tex
% \vspace{-0.5em}
In machine learning, regression tasks predict a continuous output based on a given input. Yet, if the ground truth (prediction target) is within a specific range, \textit{e.g.,} in the case of age estimation~\cite{levi2015age}, one can quantize the ground truth and cast regression into a classification problem. We refer to these techniques as Classification Approaches for Regression (CAR), and in this paper we explore CAR techniques applied on monocular depth estimation.

% MDE and Deep learning
Monocular depth estimation (MDE), which is an ill-posed problem~\cite{sinha1993recovering}, consists in predicting the scene depth given only an RGB image as the input. Deep Neural Networks (DNNs) learn the mapping between the single RGB images and their corresponding depth maps to solve MDE, %. They can obtain the mechanism of the depth representation implicitly, 
and show good performance on indoor and outdoor benchmarks~\cite{Uhrig2017THREEDV, Silberman:ECCV12}. 

% MDE and CAR
Classification Approaches for Regression (CAR) \cite{cao2017estimating, li2018monocular, fu2018deep, diaz2019soft, yang2019inferring, bhat2021adabins} have emerged recently in the spotlight among MDE algorithms. The core idea is to transfer regression to a classification problem using quantization (or discretization) strategies.
% MDE+  CAR + Uncertainty
The classification models can natively provide the confidence for prediction results, which also has the potential to improve the prediction accuracy~\cite{diaz2019soft}. DNNs are prone to two kinds of uncertainty: aleatoric uncertainty and epistemic uncertainty~\cite{kendall2017uncertainties}. It is crucial to study the uncertainty of DNNs if we want to rely on their predictions. Some works proposed to estimate the uncertainty of MDE DNNs by using an auxiliary network~\cite{Poggi_CVPR_2020, yu21bmvc}, or ensembling~\cite{2017simple}. Here we gain access to the uncertainty directly using the CAR DNN.

% contributions
This work will investigate and show the complete picture of the CAR MDE methods. The contributions are as follows:\\
% \begin{itemize}
1. We systematically summarize and formalize the all major CAR MDE mechanisms to the best of our knowledge;\\
2. We implement these mechanisms on top of two different backbones, comparing depth prediction and uncertainty quality on various evaluation metrics;\\
3. \Xuanlong{We propose a new, effective uncertainty estimation method named Expectation of Distance for CAR MDE models.}
% \end{itemize}

%% file: 3survey_method.tex
\vspace{-0.5em}
\subsection{Taxonomy of CAR MDE}\label{sec:decom}
\vspace{-0.5em}
\begin{table}
\renewcommand{\captionfont}{\footnotesize} 
\centering
\scalebox{0.50}{
\begin{tabular}{|c|c|c|c|c|c|c|} 
\hline
\rowcolor[rgb]{0.753,0.753,0.753} \multicolumn{2}{|c|}{{\cellcolor[rgb]{0.602,0.602,0.602}}} & \multicolumn{4}{c|}{\textbf{Discretization}} & {\cellcolor[rgb]{0.753,0.753,0.753}} \\
\multicolumn{2}{|c|}{{\cellcolor[rgb]{0.602,0.602,0.602}}} & \multicolumn{3}{c|}{Fully Handcrafted} & \multirow{3}{*}{Adaptive} & {\cellcolor[rgb]{0.753,0.753,0.753}} \\ 
\hhline{|>{\arrayrulecolor[rgb]{0.602,0.602,0.602}}-->{\arrayrulecolor{black}}---~>{\arrayrulecolor[rgb]{0.753,0.753,0.753}}->{\arrayrulecolor{black}}|}
\multicolumn{2}{|c|}{{\cellcolor[rgb]{0.602,0.602,0.602}}} & Handcrafted & Handcrafted & Handcrafted &  & {\cellcolor[rgb]{0.753,0.753,0.753}} \\
\multicolumn{2}{|c|}{\multirow{-4}{*}{{\cellcolor[rgb]{0.602,0.602,0.602}}\textbf{CAR-MDEs}}} & + One-hot & + Ordinal & + Smooth &  & \multirow{-4}{*}{{\cellcolor[rgb]{0.753,0.753,0.753}}\begin{tabular}[c]{@{}>{\cellcolor[rgb]{0.753,0.753,0.753}}c@{}}\textbf{Post}\\\textbf{processing}\end{tabular}} \\
\hline
\cellcolor[rgb]{0.753,0.753,0.753}& \multirow{2}{*}{\begin{tabular}[c]{@{}c@{}}\small KL divergence/\\\small Weighted CE loss\end{tabular}} & - & - & SORN~\cite{diaz2019soft} & - & Argmax \\ 
\cline{3-7} {\cellcolor[rgb]{0.753,0.753,0.753}}
 & - & - & - & Cao et al.~\cite{cao2017estimating} (\cite{yin2019enforcing}) & - & \multirow{4}{*}{\begin{tabular}[c]{@{}c@{}}Soft\\weighted\\sum\end{tabular}} \\ 
\cline{2-6} {\cellcolor[rgb]{0.753,0.753,0.753}}
 & \small CE loss &{\begin{tabular}[c]{@{}c@{}}Li et al.~\cite{li2018monocular} (\cite{li2018deep, liebel2019multidepth})\end{tabular}}& - & - & - &  \\ 
\cline{2-6} {\cellcolor[rgb]{0.753,0.753,0.753}}
 & \small Multiple BCE loss & - & - & Yang et al.~\cite{yang2019inferring} & - &  \\ 
\cline{2-6} {\cellcolor[rgb]{0.753,0.753,0.753}}
 & \small Regression loss & DS-SIDENet~\cite{ren2019deep} & - & - & {\begin{tabular}[c]{@{}c@{}} Adabins~\cite{bhat2021adabins}\end{tabular}} &  \\ 
\cline{2-7} {\cellcolor[rgb]{0.753,0.753,0.753}}\multirow{-6}{*}{\rotr{\textbf{Loss function}}}
 & {\begin{tabular}[c]{@{}c@{}}Ordinal\\Regression loss\end{tabular}} & - & {\begin{tabular}[c]{@{}c@{}}DORN~\cite{fu2018deep}\\(\cite{lo2021depth, phan2021ordinal})\end{tabular}} & - & - & Ordinal sum \\
\hline
\end{tabular}
}
\vspace{-0.5em}
\caption{Summary on CAR-MDE solutions. In parentheses are methods that use the corresponding schemes as part of their solutions. The categories under Discretization, Loss function and Post processing are detailed in Sec.\ref{sec:survey_method}.}
\label{tab:three_keys}
\end{table}
To better unify the terms and make it easier to grasp the differences in contributions of CAR strategies, we propose to decompose the CAR problems into three key components: \textit{discretization}, \textit{loss function} and \textit{post-processing}. Table~\ref{tab:three_keys} offers an overview of the specific strategies used in the previous works. The details are provided in the following sections. 

The contributions of previous CAR based MDE solutions fall in two main groups: 1. novel strategies in the three components mentioned above~\cite{cao2017estimating, li2018monocular, fu2018deep, diaz2019soft, yang2019inferring, bhat2021adabins}; 2. architecture and/or loss modifications based on the previous strategies~\cite{li2018deep, lo2021depth, liebel2019multidepth, phan2021ordinal, yin2019enforcing, ren2019deep}. In most papers, CAR can improve model accuracy, making it outperform its regression version~\cite{cao2017estimating, li2018monocular, fu2018deep, diaz2019soft, yang2019inferring, bhat2021adabins, li2018deep}, or may improve model performance as part of multi-task learning~\cite{ren2019deep, liebel2019multidepth}.
\vspace{-1em}
\subsubsection{\textbf{General notations}}\label{sec:notations}

Let us first consider a monocular depth dataset $D=\{(\mathbf{x}_i, \mathbf{d}_i)\}_i$, where $\mathbf{x}_i\in{\mathbb{R}^{3\times H\times W}}$, and $\mathbf{d}_i\in{(\mathbb{R}^{+})^{H\times W}}$ represents the ground truth depth $\mathbf{d}_i$ for the image $\mathbf{x}_i$. We denote $\{d_{i,j}\}_j^N$ all the pixel values in $\mathbf{d}_i$, where $N$ is the number of pixels with valid ground truth. $a,b$ are two real values representing the minimum and maximum depth value for the dataset.
%and $d_{i,j} \in [a, b]$ where $[a, b]$ is the min-max depth value among $\{\mathbf{d}_i\}_i$. 

For CAR strategies, we denote $K$ the number of classes, which represents the level of discretization. Additionally, to simplify the notations, we use $\log$ as the logarithm with base $e$ for all papers except for~\cite{li2018deep, ren2019deep}, where the $\log$ refers to the logarithm with base $10$.
We denote  $f_{\theta_1}$  the DNN with parameters $\theta_1$.
%Regarding to the DNN, we denote the backbone $f$ with parameters $\theta_1$. 
Given $\mathbf{x}_i$, the prediction of $f_{\theta_1}$:
{\footnotesize
\begin{equation}
\hat{\mathbf{y}}_{i} = \frac{e^{ f_{\theta_1}(\mathbf{x}_i)} }{\sum^{c-1}_{p=0} e^{ [f_{\theta_1}(\mathbf{x}_i)]_{p} }}%\hspace{1em}\mathbf{l}_i= f_{\theta_1}(\mathbf{x}_i)
\end{equation}
}%
% \end{footnotesize}
where $f_{\theta_1}(\mathbf{x}_i) \in \mathbb{R}^{c \times H \times W}$ is the logit map and $[f_{\theta_1}(\mathbf{x}_i)]_{p}$ its $p$-th \Xuanlong{coefficient}, and $\hat{\mathbf{y}}_{i}$ is the Softmax output. The number of its channels $c = K$ by default, otherwise equals to the specific settings as in DORN~\cite{fu2018deep} ($2K$) and Adabins~\cite{bhat2021adabins} (128).

% %\vspace{-0.5em}
\subsubsection{\textbf{Discretization: Fully Handcrafted}}
The discretization function will output two components given $\mathbf{d}$: a \textit{depth table} $\Bar{\mathbf{d}} = \{\Bar{d_p}\}^K_p \in \mathbb{R}^{K}$ where the possible discrete depth values are set ordinally, and an indicator map equivalent to a \textit{classification map} $\mathbf{y}_i = \{{\{{y_{i,j,p}}\}_p^K}\}_j^N \in {(\mathbb{R}^{+})}^{K\times H\times W}$ which points for each pixel the closest discrete depth value. This closest depth value can be considered as a class, leading to a classification task.
Both $\Bar{\mathbf{d}}$ and $\mathbf{y}_i$ are \textit{handcrafted}, and the goal becomes the learning of $\mathbf{y}_i$.

\textbf{Handcrafted $\Bar{\mathbf{d}}$}: $\Bar{\mathbf{d}}$ contains $K$ values representing the centers of intervals $[\Bar{\Bar{d}}_0, \Bar{\Bar{d}}_1[ , .. , [\Bar{\Bar{d}}_{K-1}, \Bar{\Bar{d}}_{K}[$ with an interval width $q$:
% \begin{footnotesize}
{\footnotesize
\begin{align}
\centering
&\Bar{\mathbf{d}} = \{\Bar{d_p}\}^K_p = \{(\Bar{\Bar{d}}_0+ \Bar{\Bar{d}}_1)/2, ..., (\Bar{\Bar{d}}_{K-1}+\Bar{\Bar{d}}_{K})/2\}\label{eq:handmade}\\
&\Bar{\Bar{d}}_k = \log a + k\cdot q, k \in [0, K], q = (\log b - \log a)/K\nonumber
\end{align}
}%
% \end{footnotesize}

\textbf{Handcrafted $\Bar{\mathbf{d}}$ + One-hot $\mathbf{y}_i$}: Given $\Bar{\mathbf{d}}$, constructing $\mathbf{y}_i$ is done using one-hot encoding, as applied in~\cite{li2018monocular, li2018deep, liebel2019multidepth, ren2019deep}:
% \begin{footnotesize}
{\footnotesize
\begin{align}\label{eq:onehot}
&\mathbf{y}^{[\text{onehot}]}_{i,j} = 
\begin{bmatrix}
y_{i,j,0} & \ldots & y_{i,j,k} &\ldots & y_{i,j,K-1}
\end{bmatrix} \in \mathbb{R}^K\\ \nonumber
&\mbox{with } y_{i,j,k} = 1, \text{ if } k = \round{(\log(d_{i,j}/a))/q} \text{, and } 0\text{ otherwise.} 
\end{align}
}%
% \end{footnotesize}
in which, $\round{.}$ is a rounding operator, $q$ is defined in Eq.~\ref{eq:handmade}.

\textbf{Handcrafted $\Bar{\mathbf{d}}_i$ + Ordinal $\mathbf{y}_i$}: Furthermore, there are several variants of Eq~\ref{eq:onehot}. Ordinal properties can be applied on it as presented in~\cite{fu2018deep} and the followed works~\cite{lo2021depth,phan2021ordinal}:
% \begin{footnotesize}
{\footnotesize
\begin{align}\label{eq:ordinal}
&\mathbf{y}^{[\text{ordi}]}_{i,j} = 
\begin{bmatrix}
y_{i,j,0} & \ldots & y_{i,j,k} &\ldots & y_{i,j,K-1}
\end{bmatrix} \in \mathbb{R}^K \\ \nonumber
&\mbox{with } y_{i,j,k} = 1, \text{ if } k \leq \round{(\log(d_{i,j}/a))/q} \text{, and } 0\text{ otherwise.} 
\end{align}
}%
% \end{footnotesize}
\textbf{Handcrafted $\Bar{\mathbf{d}}_i$ + Smooth $\mathbf{y}_i$}:
It is also possible to predict a smooth discrete map from the initial discrete map $\mathbf{y}^{[\text{onehot}]}_{i,j}$. The indicator in the classification map is softened by applying on  $\Bar{\mathbf{d}}$ a Gaussian kernel, %like in kernel density estimation 
as to predict distance within a coarser range. %This discrete distribution will make the vector in Eq~\ref{eq:onehot} smoother:
The smooth $\mathbf{y}_i$ are defined by:
% \begin{footnotesize}
{\footnotesize
\begin{align}
\centering
    &\mathbf{y}{^{[\text{smo1}]}_{i,j}} = e^{- \gamma || \log(d_{i,j}) - \Bar{\mathbf{d}}||^2} \label{eq:softlabel1}\\
    &\mathbf{y}{^{[\text{smo2}]}_{i,j}} = \frac{e^{- \gamma || \log(d_{i,j}) - \Bar{\mathbf{d}}||^2}}{\sum_{p=0}^{K-1} {e^{- \gamma || \log(d_{i,j}) - {\Bar{d}}_p||^2}}} \label{eq:softlabel2}
\end{align}
}%
% \end{footnotesize}
where $\gamma$ is a hyperparameter which can be regarded as the scale of the discrete distribution (the smaller $\gamma$, the flatter the label distribution in $\mathbf{y}{_{i,j}}$). Specifically, Yang et al.~\cite{yang2019inferring} use Eq~\ref{eq:softlabel1} as the unnormalized soft target labels, while SORN~\cite{diaz2019soft} applies the normalized version in Eq~\ref{eq:softlabel2}. \Xuanlong{Moreover, Cao et al.~\cite{cao2017estimating} introduce a $K \times K$ symmetric ``information gain" matrix $H$ in their loss function with elements $H(k,\mathbf{p}) = e^{-\gamma||k-\mathbf{p}||^2}$, where $\mathbf{p} = [0,...,K-1]$ and $k$ is the discrete ground truth index as defined in Eq~\ref{eq:onehot}. In this case:
{\footnotesize
\begin{equation}\label{eq:smooth3}
\mathbf{y}^{\text{[smo3]}}_{i,j} = 
e^{-\gamma||k-\mathbf{p}||^2} = e^{- \gamma\cdot q^{-2} ||\log(d_{i,j}) - (\Bar{\mathbf{d}} - 0.5q)||^2}    
\end{equation}
}%
Since $q$ is a constant, this strategy can be regarded as being equivalent to the $\mathbf{y}^{\text{[smo1]}}_{i,j}$ in Eq~\ref{eq:softlabel1}.}
%\vspace{-1.5em}
\subsubsection{\textbf{Discretization: Adaptive}}
% %\vspace{-0.5em}
In the absence of the handcrafted depth table or the classification map, one may also implicitly train both of them using a regression loss as in Adabins~\cite{bhat2021adabins}. Thus the goal of the DNN is changed from fitting the handcrafted classification maps to fitting the continuous ground truth depth, while still following the principle of building depth tables and classification maps.

In Adabins~\cite{bhat2021adabins}, the depth table is implicitly trained along with the classification map using a non-linear block %\footnote{which is a DNN}
$g_{\theta_2}$ with $\theta_2$ the parameters of $g$,  which is a mini ViT ~\cite{bhat2021adabins}. In this case, $g_{\theta_2}$ is set on top of the backbone $f_{\theta_1}$, and it will output $\hat{\Bar{\mathbf{d}}}_i^{[\text{ada}]}$ and $\hat{\mathbf{y}}_i^{[\text{ada}]}$ given $f_{\theta_1}(\mathbf{x}_i)$: %$\hat{\mathbf{y}}^{\prime [\theta_1]}_i$.
% \begin{footnotesize}
{\footnotesize
\begin{align}
&\hat{\Bar{\mathbf{d}}}^{[\text{ada}]}_i = \{a + (b-a)(\sum_{s=0}^{p}{\hat{\Bar{d}}^{\text{[ada]}}_{i,s}})\}_{p=0}^{K-1},\hat{\mathbf{y}}_{i,j}^{[\text{ada}]} = \frac{e^{\mathbf{l}_{i,j} } }{\sum^{K-1}_{p=0} e^{l_{i,j,p}  }}\label{eq:learned2}\\
&\text{with }\{\hat{\Bar{d}}^{\text{[ada]}}_i\}_p^K, \mathbf{l}_i = g_{\theta_2}(f_{\theta_1}(\mathbf{x}_i) )\nonumber
\end{align}
}%
%\begin{align}
%&\hat{\Bar{\mathbf{d}}}^{[\theta_1,\theta_2]}_i = \{a + (b-a)({\hat{\Bar{\Bar{d}}}^{\prime [\theta_1, \theta_2]}_{i,p}} + \sum_{s=0}^{p-1}{\hat{\Bar{\Bar{d}}}^{\prime [\theta_1, \theta_2]}_{i,s}})\}_{p=0}^{K-1}\label{eq:learned1}\\
%&\hat{\Bar{\mathbf{d}}}_i^{\prime [\theta_1,\theta_2]} = \frac{\text{ReLU}(\hat{\Bar{\mathbf{d}}}_i^{\prime})}{\sum_{p\text{=}0}^{K\text{-}1}{{\text{ReLU}(\hat{\Bar{\Bar{d^{\prime}}}}_{i,p}}})}, \hat{\mathbf{y}}_{i,j}^{[\theta_1,\theta_2]} = \frac{e^{\hat{\mathbf{y}}^{\prime [\theta_1,\theta_2]}_{i,j}}}{\sum^{K-1}_{p=0} e^{\hat{y}^{\prime [\theta_1,\theta_2]}_{i,j,p}}};\label{eq:learned2}\\
%&\hat{\Bar{\mathbf{d}}}_i^{\prime [\theta_1,\theta_2]}, \hat{\mathbf{y}}_i^{\prime [\theta_1,\theta_2]} = g(\hat{\mathbf{y}}^{\prime [\theta_1]}_i; \theta_2)\label{eq:learned3}
%% \\& \hat{\mathbf{y}}_{i,j}^{\text{[f]}}(\theta) = \frac{e^{\hat{\mathbf{y}}^{\prime [f]}_{i,j}}}{\sum^{K\text{-}1}_{p\text{=}0} e^{\hat{y}^{\prime [f]}_{i,j,p}}},
%% \hat{\mathbf{y}}_i^{\prime [f]}(\theta) = f(\mathbf{x}_i; \theta)\label{eq:learn2}
%\end{align}
% \end{array}
% \end{equation}
% \end{footnotesize}
where, $\hat{\mathbf{y}}_{i,j}^{[\text{ada}]}$ is the product of a Softmax function, and $\hat{\Bar{\mathbf{d}}}_i^{[\text{ada}]}$ is a cumulative summation output followed by a normalization operation which is %omitted and 
included in $g_{\theta_2}$. Since $\hat{\Bar{\mathbf{d}}}_i^{[\text{ada}]}$ is a product of $g_{\theta_2}$ taking $f_{\theta_1}(\mathbf{x}_i)$, for each $\mathbf{x}_i$, not only a unique classification map but also a unique depth table will be provided. 

%\vspace{-2em}
\subsubsection{\textbf{Loss function}}
%\vspace{-0.5em}
Based on the previous discretization strategies, we introduce here the loss function design. Models should fit their output to the designed $\mathbf{y}$ or $\mathbf{d}$. For brevity, we define first the total loss $L_{\text{total}} = \sum_{i}\sum_{j=0}^{N-1}L_{{i,j}}$ where $L_{i,j}$ is the loss for pixel $j$, on the $i$-th data. We just define $L_{i,j}$ in the following sections for simplicity.

\textbf{Cross entropy (CE) loss}: is a straightforward solution given an one-hot classification map: ${L_{i,j}^{{[\text{CE}]}}}(\theta_1) = -({{\mathbf{y}^{[\text{onehot}]}_{i,j}} \log \hat{\mathbf{y}}_{i,j}})$.

\textbf{Ordinal regression loss}: is essentially an implicit ordinal selection plus a multiple binary cross entropy (BCE) loss. Instead of directly using $\hat{\mathbf{y}}_{i,j}$, it requires to do an ordinal selection on the logit map $f_{\theta_1}(\mathbf{x}_i)$ with $c = 2K$ to $c=K$ as the predicted classification map, then to apply a Multiple-BCE loss on it:
% \begin{footnotesize}
{\footnotesize
\begin{align}\label{eq:ordinalloss}
&{L_{i,j}^{{[\text{ordi}]}}}(\theta_1) \text{=} -[{{\mathbf{y}^{[\text{ordi}]}_{i,j}} \log \hat{\mathbf{y}}_{i,j}^{{[\text{ordi}]}} + (1 - {\mathbf{y}^{[\text{ordi}]}_{i,j}}) \log (1 - \hat{\mathbf{y}}_{i,j}^{{[\text{ordi}]}})}]\\
&\text{with }\hat{\mathbf{y}}_{i,j}^{{[\text{ordi}]}} = \frac{e^{[f_{\theta_1}(\mathbf{x}_i)]_{2p+1}}}{e^{[f_{\theta_1}(\mathbf{x}_i)]_{2p+1}} + e^{[f_{\theta_1}(\mathbf{x}_i)]_{2p}}}\nonumber
\end{align}
}%
% \end{footnotesize}
\Xuanlong{where $2p+1$ and $2p$ represent the indices of the coefficient.}

\textbf{Weighted CE loss}: is applied when the target vector is a soft discrete distribution. The CE loss turns to be equal to: ${L_{i,j}^{{[\text{WCE}]}}}(\theta_1) = - ({\mathbf{y}{^{[\text{smo2}]}_{i,j}} \log \hat{\mathbf{y}}_{i,j}})$,
and it has the same form as the Kullback-Leibler divergence loss.

\textbf{Multiple BCE loss}: is another solution when the target is a soft discrete distribution. Yang et al.~\cite{yang2019inferring} apply BCE loss on every class value in $\mathbf{y}^{[\text{smo1}]}_{i,j}$ defined in Eq.~\ref{eq:softlabel1}.

\textbf{Regression losses}:
DS-SIDENet~\cite{ren2019deep} applies CAR with a smooth L1 loss~\cite{chang2018pyramid} to fit the one-hot classification map target. Conversely, Adabins~\cite{bhat2021adabins} combines the per-image adaptive depth table $\hat{\Bar{\mathbf{d}}}^{[\text{ada}]}_i$ and $\hat{\mathbf{y}}_{i,j}^{[\text{ada}]}$ defined in Eq.~\ref{eq:learned2} as the predicted depth, then applies a Scale-Invariant loss~\cite{eigen2014depth}.

\subsubsection{\textbf{Post-processing}}\label{sec:Post-processing}
Post-processing aims to restore the discrete predicted labels to continuous depth values. In the following equations, we use the power function in base $e$, see Sec.~\ref{sec:notations}.

\textbf{Ordinal sum}: For DORN~\cite{fu2018deep}, the continuous depth is restored from the sum of the output Sigmoid labels which are higher than  or equal to 0.5:
{\footnotesize
\begin{equation}
    % \hat{d}_{i,j} = e^{\log(a) + q \cdot \left[\sum_{p=0}^{K-1}\mathbbm{1}\{\hat{y}^{\text{[ordi]}}_{i,j,p}\geq 0.5\} + 0.5\right]}
    \hat{d}_{i,j} = \text{exp}\{{\log(a) + q \cdot [\sum_{p=0}^{K-1}\mathbbm{1}\{\hat{y}^{\text{[ordi]}}_{i,j,p}\geq 0.5\} + 0.5]}\}
    \label{eq:restoreordinal}
\end{equation}
}%
where $q$ and $\hat{y}^{\text{[ordi]}}_{i,j,p}$ are defined in Eq.~\ref{eq:handmade} and Eq.~\ref{eq:ordinalloss} respectively.

\textbf{Soft weighted sum}: is a solution that may applied on both handcrafted or learned depth tables. It sums the Hadamard product between the depth table and the classification map:
{\footnotesize
\begin{align}\label{eq:softweightedsum1}
    % \hat{d}_{i,j} = e ^{\sum_{p=0}^{K-1}{\Bar{d}_{p} \cdot \hat{y}_{i,j,p}}}
\hat{d}_{i,j} = \text{exp}\{\sum_{p=0}^{K-1}{\Bar{d}_{p} \cdot \hat{y}_{i,j,p}}\}
\end{align}
}%
Note that essentially AdaBins~\cite{bhat2021adabins} also follows this pattern.

\textbf{Argmax}: The authors of SORN~\cite{diaz2019soft} claim that Argmax outperforms Soft weighted sum in their case:
{\footnotesize
\begin{equation}
    % \hat{d}_{i,j} = e^{\log(a) + q \cdot [ \underset{p}{\mathrm{argmax}}\,(\{\hat{y}_{i,j,p}\}^K_p) + 0.5]}\label{eq:argmax}
       \hat{d}_{i,j} = \text{exp}\{\log(a) + q \cdot [ \underset{p}{\mathrm{argmax}}\,(\{\hat{y}_{i,j,p}\}^K_p) + 0.5]\}\label{eq:argmax}
\end{equation}
}%
where $q$ is defined in Eq.~\ref{eq:handmade}.
%\vspace{-1em}

\subsection{Uncertainty estimation of CAR MDE}\label{sec:uncertaintyCAR}
%\vspace{-0.5em}
In this section, we will discuss the previous works on uncertainty estimation for CAR MDE, the difficulty of this problem and our proposed approaches on estimating CAR uncertainty.

%MDE is a dense regression problem, and t
The ground truth uncertainty or the Oracle should be the model's prediction error. The previous works on MDE uncertainty estimation~\cite{yu21bmvc, Poggi_CVPR_2020} mainly use the principle of learning the prediction error~\cite{kendall2017uncertainties}. Meanwhile, the \textbf{Variance} among the point estimations given by MC-Dropout~\cite{gal2016dropout} and Deep Ensembles~\cite{2017simple} can also be applied for this task. %Unlike the previous works, the likelihoods or weights in the predicted classification map in CAR can offer another possibility on estimating the uncertainty which is mentioned in the previous works but rarely discussed. 
Unlike the previous works, the likelihoods of the predicted class (the quantified depth value) given the input data provided by CARs can offer another possibility to estimate the uncertainty mentioned in the previous works but rarely discussed. 
Yang et al.~\cite{yang2019inferring} suggest to use \textbf{Shannon Entropy} (\textbf{S-Entr})~\cite{shannon2001mathematical} among the output Softmax classification map:
$-\sum_{p=0}^{K-1}\hat{{y}}_{i,j,p} \log{\hat{{y}}_{i,j,p}}$
Moreover, they showed cases where the depth is well predicted, yet the entropy is high, leading to an under-confident uncertainty score.
Other strategies such as \textbf{1-Maximum Class Probability} (\textbf{1-MCP}) can also be regarded as the uncertainty: $1-\max_p(\{\hat{{y}}_{i,j,p}\}_p)$
These are typical solutions used in classification tasks, and we argue that they will be suitable in case of using Argmax in post-processing for CAR problems. Widely used soft weighted sum (see Table~\ref{tab:three_keys}) makes the property of CAR special: not only the classification map but also the depth table should be taken into account in the final result as shown in Eq.~\ref{eq:softweightedsum1}.

Following these remarks, we propose a new solution for CAR uncertainty. We first note that the previously mentioned methods lack consideration of the depth table, and further its relationship to the classification map. Hence, we define as CAR uncertainty metric: the \textbf{Expectation of Distance} (\textbf{E-Dist}) between the quantified depth values (either handcrafted logarithm depth table $\Bar{\mathbf{d}}$ (Eq.~\ref{eq:handmade}) or the learned one $\hat{\Bar{\mathbf{d}}}^{\text{[ada]}}_i$ (Eq.~\ref{eq:learned2})) and the final predicted depth $\hat{\mathbf{d}}_i$ (Eq.~\ref{eq:softweightedsum1},~\ref{eq:argmax}) :
{\footnotesize
\begin{equation}\label{eq:e-dist}
    \text{E-Dist} = \sum_{p=0}^{K-1}{\hat{{y}}_{i,j,p} \cdot (e^{\Bar{d}_{p}} - \hat{d}_{i,j})^2} \hspace{0.5em}\text{or} \hspace{0.5em}
    \text{E-Dist} = \sum_{p=0}^{K-1}{\hat{{y}}_{i,j,p} \cdot (\hat{\Bar{d}}^{\text{[ada]}}_{i,p} - \hat{d}_{i,j})^2}
\end{equation}
}%

Additionally, to our knowledge, no previous works discuss the uncertainty of ordinal regression model~\cite{fu2018deep}. According to its CAR strategy, only values greater than or equal to 0.5 in its classification map will be considered in the final depth calculation, thus we argue that the uncertainty comes from this part. The modified E-Dist for ordinal regression is as below: We propose to discretize the depth prediction $\hat{\mathbf{d}}_i$ (Eq.~\ref{eq:restoreordinal}) using Eq.~\ref{eq:ordinal}, that we denote as ${\mathbf{y}^{\prime[\text{ordi}]}_{i,j}}$. Then we calculate the distance between ${\mathbf{y}^{\prime[\text{ordi}]}_{i,j}}$ and ${\hat{\mathbf{y}}^{[\text{ordi}]}_{i,j}}$ (defined in Eq.\ref{eq:ordinalloss}) weighted by the depth table and only consider the part with $\hat{\mathbf{y}}^{\text{[ordi]}}_{i,j}\geq 0.5$:
% \begin{footnotesize}
{\footnotesize
\begin{equation}
    \sum_{p=0}^{K-1}{e^{\Bar{d}_{p}} \cdot ({y^{\prime[\text{ordi}]}_{i,j,p}} - {\hat{y}^{[\text{ordi}]}_{i,j,p}})^2} \cdot \mathbbm{1}\{\hat{y}^{\text{[ordi]}}_{i,j,p}\geq 0.5\}
\end{equation}
}%

%% file: 4experiments.tex
\begin{table*}[t!]
\renewcommand{\captionlabelfont}{\bf}
\renewcommand{\captionfont}{\footnotesize} 
    \centering
    \scalebox{0.85}{
    \subfloat[][\label{tab:fcn_depth_kitti}]{%
        \resizebox{1.2\columnwidth}{!}{%
\begin{tabular}{lcccccccc||cccccccc||c}
\toprule
\textbf{Backbones} & \multicolumn{8}{c}{\textbf{BTS}} & \multicolumn{8}{c}{\textbf{FCN}} & \multirow{3}{*}{\begin{tabular}[c]{@{}l@{}}$\textbf{K}$\\\end{tabular}}  \\
 \cmidrule(lr){2-9}\cmidrule(lr){10-17}
\textbf{Metrics} & {\cellcolor[rgb]{0.58,0.737,0.914}}\textbf{$\delta$1$\uparrow$} & {\cellcolor[rgb]{0.58,0.737,0.914}}\textbf{$\delta$2$\uparrow$} & {\cellcolor[rgb]{0.58,0.737,0.914}}\textbf{$\delta$3$\uparrow$} & \begin{tabular}[c]{@{}c@{}}{\cellcolor[rgb]{0.914,0.631,0.58}}\textbf{Abs}\\{\cellcolor[rgb]{0.914,0.631,0.58}}\textbf{Rel\textbf{$\downarrow$}}\end{tabular} & \begin{tabular}[c]{@{}c@{}}{\cellcolor[rgb]{0.914,0.631,0.58}}\textbf{Sq}\\{\cellcolor[rgb]{0.914,0.631,0.58}}\textbf{Rel\textbf{$\downarrow$}}\end{tabular} & {\cellcolor[rgb]{0.914,0.631,0.58}}\textbf{RMSE$\downarrow$} & \begin{tabular}[c]{@{}c@{}}{\cellcolor[rgb]{0.914,0.631,0.58}}\textbf{RMSE}\\{\cellcolor[rgb]{0.914,0.631,0.58}}\textbf{log\textbf{$\downarrow$}}\end{tabular} & {\cellcolor[rgb]{0.914,0.631,0.58}}\textbf{log10$\downarrow$} & \multicolumn{1}{c}{{\cellcolor[rgb]{0.58,0.737,0.914}}\textbf{$\delta$1$\uparrow$}} & \multicolumn{1}{c}{{\cellcolor[rgb]{0.58,0.737,0.914}}\textbf{$\delta$2$\uparrow$}} & \multicolumn{1}{c}{{\cellcolor[rgb]{0.58,0.737,0.914}}\textbf{$\delta$3$\uparrow$}} & \multicolumn{1}{c}{\begin{tabular}[c]{@{}c@{}}{\cellcolor[rgb]{0.914,0.631,0.58}}\textbf{Abs}\\{\cellcolor[rgb]{0.914,0.631,0.58}}\textbf{Rel\textbf{$\downarrow$}}\end{tabular}} & \multicolumn{1}{c}{\begin{tabular}[c]{@{}c@{}}{\cellcolor[rgb]{0.914,0.631,0.58}}\textbf{Sq}\\{\cellcolor[rgb]{0.914,0.631,0.58}}\textbf{Rel\textbf{$\downarrow$}}\end{tabular}} & \multicolumn{1}{c}{{\cellcolor[rgb]{0.914,0.631,0.58}}\textbf{RMSE$\downarrow$}} & \multicolumn{1}{c}{\begin{tabular}[c]{@{}c@{}}{\cellcolor[rgb]{0.914,0.631,0.58}}\textbf{RMSE}\\{\cellcolor[rgb]{0.914,0.631,0.58}}\textbf{log\textbf{$\downarrow$}}\end{tabular}} & \multicolumn{1}{c||}{{\cellcolor[rgb]{0.914,0.631,0.58}}\textbf{log10$\downarrow$}}\\ 
\toprule
DORN~\cite{fu2018deep} & \first 0.952 & \first 0.992 & \first 0.998 & \second 0.069 & \first 0.267 & \first 2.802 & \first 0.103 & \first 0.029 & \first 0.940 & \second 0.990 & \first 0.998 & 0.076 & \first 0.292 & \first 2.962 & \first 0.113 & \second 0.033 & 80\\ 
% \midrule
Cao et al.~\cite{cao2017estimating} & 0.945 & \first 0.992 & \first 0.998 & 0.077 & 0.292 & 2.988 & 0.111 & 0.034 & 0.928 & 0.989 & \first 0.998 & 0.084 & 0.344 & 3.223 & 0.122 & 0.036 & 50\\ 
% \midrule
Li et al.~\cite{li2018monocular} & 0.950 & 0.990 & \first 0.998 & 0.070 & 0.287 & 2.928 & \second 0.106 & \second 0.030 & \first 0.940 & 0.988 & \second 0.997 & \second 0.075 & 0.314 & 3.190 & \second 0.116 & \second 0.033 & 150~\cite{li2018deep}\\ 
% \midrule
SORN~\cite{diaz2019soft} & 0.947 & \first 0.992 & \first 0.998 & 0.071 & 0.290 & 2.929 & 0.107 & 0.031 & 0.863 & 0.976 & 0.995 & 0.119 & 0.563 & 3.938 & 0.163 & 0.051 & 120\\ 
% \midrule
Yang et al.~\cite{yang2019inferring} & \second 0.951 & \second 0.991 & \first 0.998 & \first 0.065 & 0.276 & 2.897 & \first 0.103 & \first 0.029 & \first 0.940 & 0.989 & \second 0.997 & \first 0.072 & \second 0.302 & 3.096 & \first 0.113 & \first 0.032 & 128\\ 
DS-SIDE~\cite{ren2019deep} & 0.950 & \second 0.991 & \first 0.998 & 0.071 & \second 0.275 & \second 2.886 & \second 0.106 & 0.032 & 0.931 & \second 0.990 & \first 0.998 & 0.079 & 0.331 & 3.353 & 0.119 & 0.035 & 80\\
% \midrule
Adabins~\cite{bhat2021adabins} & 0.935 & 0.990 & \first 0.998 & 0.078 & 0.347 & 3.143 & 0.114 & 0.033 & \second 0.937 & \first 0.991 & \first 0.998 & 0.079 & 0.331 & \second 3.027 & \first 0.113 & \second 0.033 & 256\\

\midrule

DORN~\cite{fu2018deep} & \second 0.952 & \second 0.992 & \first 0.998 & \second 0.069 & \first 0.267 & \first 2.802 & \first 0.103 & \first 0.029 & \first 0.940 & \first 0.990 & \first 0.998 & \first 0.076 & \first 0.292 & \first 2.962 & \second 0.113 & \second 0.033 & 80\\ 
% \midrule
Cao et al.~\cite{cao2017estimating} & \first 0.953 & 0.991 & \first 0.998 & \first 0.066 & \second 0.268 & \second 2.857 & \first 0.103 & \first 0.029 & 0.934 & \second 0.989 & \second 0.997 & \first 0.076 & 0.319 & 3.124 & 0.117 & \second 0.033 & 80\\ 
% \midrule
Li et al.~\cite{li2018monocular} & 0.949 & 0.990 & \second 0.997 & 0.087 & 0.305 & 2.982 & 0.116 & 0.037 & 0.933 & 0.988 & \second 0.997 & 0.096 & 0.350 & 3.157 & 0.126 & 0.040 & 80\\ 
% \midrule
SORN~\cite{diaz2019soft} & 0.949 & \first 0.993 & \first 0.998 & 0.072 & 0.283 & 2.902 & \second 0.106 & \second 0.031 & 0.863 & 0.976 & 0.995 & 0.122 & 0.573 & 3.950 & 0.165 & 0.052 & 80\\ 
% \midrule
Yang et al.~\cite{yang2019inferring} & 0.948 & 0.991 & \first 0.998 & 0.070 & 0.284 & 2.973 & 0.107 & \second 0.031 & \first 0.940 & \first 0.990 & \second 0.997 & \first 0.076 & \second 0.308 & 3.067 & 0.115 & 0.034 & 80\\ 
DS-SIDE~\cite{ren2019deep} & 0.950 & 0.991 & \first 0.998 & 0.071 & 0.275 & 2.886 & \second 0.106 & 0.032 & 0.931 & \first 0.990 & \first 0.998 & \second 0.079 & 0.331 & 3.353 & 0.119 & 0.035 & 80\\ 
% \midrule
Adabins~\cite{bhat2021adabins} & 0.933 & 0.989 & \first 0.998 & 0.079 & 0.357 & 3.203 & 0.116 & 0.033 & \second 0.937 & \first 0.990 & \first 0.998 & \first 0.076 & 0.318 & \second 3.062 & \first 0.112 & \first 0.032 & 80\\
\midrule
Org & 0.955 & 0.993 & 0.998 & 0.060 & 0.249 & 2.798 & 0.096 & 0.027 & 0.944 & 0.992 & 0.998 & 0.069 & 0.275 & 2.938 & 0.107 & 0.030 & 1\\ 
% \midrule
MC-Dropout~\cite{gal2016dropout} & 0.941 & 0.992 & 0.998 & 0.083 & 0.308 & 2.910 & 0.114 & 0.035 & 0.918 & 0.984 & 0.996 & 0.085 & 0.369 & 3.157 & 0.125 & 0.036 & 1\\ 
% \midrule
Deep Ensembles~\cite{2017simple} & 0.957 & 0.993 & 0.999 & 0.059 & 0.233 & 2.688 & 0.093 & 0.026 & 0.946 & 0.992 & 0.998 & 0.068 & 0.269 & 2.923 & 0.106 & 0.030 & 1\\ 
\bottomrule
\end{tabular}
        }
    }
        \hspace{1em}
    \subfloat[][\label{tab:bts_depth_kitti}]{%
        \resizebox{1.03\columnwidth}{!}{%
\begin{tabular}{lcccccc||cccccc||c} 
\toprule
\textbf{Backbones} & \multicolumn{6}{c}{\textbf{BTS}} & \multicolumn{6}{c}{\textbf{FCN}} &
\multirow{3}{*}{\begin{tabular}[c]{@{}l@{}}$\textbf{K}$\\\end{tabular}}\\
\cmidrule(lr){2-7}\cmidrule(lr){8-13}
\textbf{Metrics} & \multicolumn{3}{c}{{\cellcolor[rgb]{0.914,0.631,0.58}}\textbf{AUSE~RMSE\textbf{$\downarrow$}}} &
\multicolumn{3}{c}{{\cellcolor[rgb]{0.914,0.631,0.58}}\textbf{AUSE~AbsRel\textbf{$\downarrow$}}} & \multicolumn{3}{c}{{\cellcolor[rgb]{0.914,0.631,0.58}}\textbf{AUSE~RMSE\textbf{$\downarrow$}}} & \multicolumn{3}{c}{{\cellcolor[rgb]{0.914,0.631,0.58}}\textbf{AUSE~AbsRel\textbf{$\downarrow$}}} \\
\cmidrule(lr){2-4}\cmidrule(lr){5-7}\cmidrule(lr){8-10}\cmidrule(lr){11-13}
\textbf{Methods} & \textbf{1-MCP} & \textbf{S-Entr} & \textbf{E-Dist} & \textbf{1-MCP} & \textbf{S-Entr} & \textbf{E-Dist} & \textbf{1-MCP} & \textbf{S-Entr} & \textbf{E-Dist} & \textbf{1-MCP} & \textbf{S-Entr} & \textbf{E-Dist} \\ 
\toprule

Cao et al.~\cite{cao2017estimating} & 0.542 & 0.770 & 0.133 & 0.382 & 0.424 & 0.411 & 0.532 & 0.701 & \second 0.127 & 0.354 & 0.375 & 0.375 & 50\\ 
% \midrule
Li et al.~\cite{li2018monocular} & 0.174 & 0.153 & 0.187 & 0.276 & 0.262 & 0.409 & 0.137 & 0.138 & 0.132 & 0.241 & 0.235 & 0.259 & 150\\ 
% \midrule
SORN~\cite{diaz2019soft} & 1.371 & 1.394 & 0.157 & 0.939 & 0.982 & 0.427 & 1.244 & 1.283 & 0.170 & 0.754 & 0.755 & 0.451 & 120\\ 
% \midrule
Yang et al.~\cite{yang2019inferring} & 0.141 & 0.145 & \first 0.094 & \second 0.232 & \first 0.219 & 0.256 & 0.142 & 0.161 & \first 0.111 & \first 0.225 & \second 0.226 & 0.247 & 128\\ 
% \midrule
DS-SIDE~\cite{ren2019deep} & 0.698 & 0.806 & 0.293 & 0.525 & 0.544 & 0.397 & 1.212 & 1.331 & 0.995 & 0.630 & 0.722 & 0.484 & 80 \\
Adabins~\cite{bhat2021adabins} & 0.855 & 0.827 & 0.179 & 0.536 & 0.527 & 0.377 & 0.984 & 0.945 & 0.191 & 0.608 & 0.589 & 0.398 & 256\\
DORN~\cite{fu2018deep} &  
0.188 & 0.158 & \second 0.128 & 0.530 & 0.430 & 0.303 & 0.202 & 0.165 & 0.135 & 0.593 & 0.445 & 0.283 & 80\\
\midrule

Cao et al.~\cite{cao2017estimating} & 0.371 & 0.476 & \second 0.119 & 0.323 & 0.329 & 0.356 & 0.349 & 0.393 & \second 0.117 & 0.284 & 0.265 & 0.308 & 80\\ 
% \midrule
Li et al.~\cite{li2018monocular} & 0.206 & 0.178 & 0.181 & 0.355 & 0.348 & 0.449 & 0.170 & 0.163 & 0.124 & 0.318 & 0.310 & 0.333 & 80 \\ 
% \midrule
SORN~\cite{diaz2019soft} & 1.367 & 1.390 & 0.157 & 0.900 & 0.941 & 0.444 & 1.228 & 1.275 & 0.175 & 0.725 & 0.737 & 0.473 & 80\\ 
% \midrule
Yang et al.~\cite{yang2019inferring} & 0.194 & 0.179 & \first 0.099 & 0.273 & \first 0.259 & \second 0.271 & 0.156 & 0.169 & \first 0.104 & \second 0.258 & \first 0.252 & 0.274 & 80\\ 
% \midrule
DS-SIDE~\cite{ren2019deep} & 0.698 & 0.806 & 0.293 & 0.525 & 0.544 & 0.397 & 1.212 & 1.331 & 0.995 & 0.630 & 0.722 & 0.484 & 80 \\
Adabins~\cite{bhat2021adabins} & 0.823 & 0.683 & 0.181 & 0.499 & 0.450 & 0.360 & 0.775 & 0.710 & 0.234 & 0.502 & 0.478 & 0.391 & 80 \\
DORN~\cite{fu2018deep} &  
0.188 & 0.158 & 0.128 & 0.530 & 0.430 & 0.303 & 0.202 & 0.165 & 0.135 & 0.593 & 0.445 & 0.283 & 80\\ 
\midrule
MC-Dropout~\cite{gal2016dropout} & \multicolumn{3}{c}{0.460 (\textbf{Variance})} & \multicolumn{3}{c||}{0.501 (\textbf{Variance})} & \multicolumn{3}{c}{0.322 (\textbf{Variance})} &  \multicolumn{3}{c||}{0.456 (\textbf{Variance})} & 1\\ 
% \midrule
Deep Ensembles~\cite{2017simple} & \multicolumn{3}{c}{0.165 (\textbf{Variance})} & \multicolumn{3}{c||}{0.261 (\textbf{Variance})} & \multicolumn{3}{c}{0.184 (\textbf{Variance})} &  
\multicolumn{3}{c||}{0.290 (\textbf{Variance})} & 1 \\ 
\bottomrule
\end{tabular}
        }
    }% 
    }
    % \addtocounter{table}{+1}
    \vspace{-1em}
    \caption{
    The best/second-best values are highlighted in dark/light blue. The results from regression-based models are provided as reference (lower rows), and we only highlight the CAR-based results.
      \textbf{(a)} Depth uncertainty evaluations. MC-Dropout and Deep Ensembles will only provide the uncertainty with one method.
      \textbf{(b)} Depth accuracy evaluations. Org: the original BTS~\cite{lee2019big} model and the regression version applied on FCN~\cite{long2015fully} model. 
    }
        \vspace{-1.8em}
\end{table*}

\begin{figure}[!t]
\centering
\renewcommand{\captionfont}{\footnotesize} 
\includegraphics[width=0.28\textwidth]{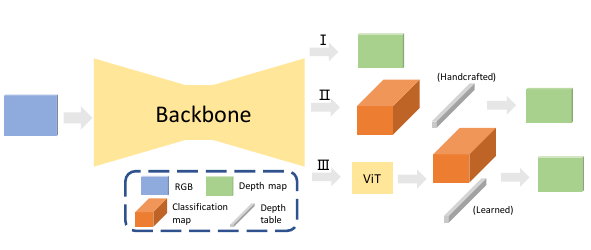}
\vspace{-0.5em}
\caption{\footnotesize Experiment pipeline. Three heads will be applied on the backbone: \textbf{I}. original regression version; \textbf{II}: MDE with handcrafted discretization; \textbf{III}: MDE with adaptive discretization through a mini ViT module~\cite{bhat2021adabins}.}
\label{fig:pipeline}
\end{figure}
\vspace{-0.5em}
In this section, we fill in the missing comparisons of the previous works. Meanwhile, our experiments provide an extensive analysis of CAR MDE uncertainty estimation. While it is not trivial to propose a model-agnostic approach, the ensuing discussion establishes some important guidelines about performing this task on CAR models. 

\vspace{-1em}
\subsection{Experiment settings}
\vspace{-0.5em}
All the experiments are based on Eigen-split~\cite{eigen2014depth} KITTI dataset~\cite{Uhrig2017THREEDV}. 
We followed the original settings in the corresponding papers for CAR strategies and applied them on a regression-based and a classification-based backbone respectively. In addition, we added experiments with $K=80$ to the methods with originally different choices for $K$ for better comparison. Fig~\ref{fig:pipeline} illustrates the experiment pipeline.\\
\textbf{Reg.-based backbone: }We use BTS-DenseNet161 \cite{lee2019big, huang2017densely}. 
% We add a classifier head as in FCN~\cite{long2015fully} for a smoother output of a multiple-channel map on the top. We find that this can get better results than just adjusting the number of output channels. 
Same BTS training settings are applied for all the methods.\\
\textbf{Cla.-based backbone: }We choose FCN-ResNet101 \cite{long2015fully, he2016deep}. FCN is originally designed for semantic segmentation, thus it is suitable for CAR methods. For the one-channel regression version (org), followed BTS, we apply a Sigmoid on the top and multiply the output by $b$.\\
\textbf{Evaluation matrices}: We use the same matrices first introduced in~\cite{eigen2014depth} and used in many subsequent works for depth performance. For uncertainty estimation, we use the area under sparsification error curve (AUSE), as in~\cite{yang2019inferring, Poggi_CVPR_2020, yu21bmvc}. 1\% of pixels are removed each time and we calculate RMSE and AbsRel for the rest. 
% The pixels are removed according to their RMSE and AbsRel from high to low as the Oracle curves, and removed based on their predicted uncertainty order as the predictive curves. The areas between the predictive curves and the Oracle curves are denoted as AUSE-RMSE and AUSE-AbsRel. 
Uncertainty estimation methods we used are introduced in Sec.~\ref{sec:uncertaintyCAR}. We will compare the CAR MDE uncertainty with widely used MC-Dropout~\cite{gal2016dropout} (with 8 forward passes) and Deep Ensembles~\cite{2017simple} (with 3 models).\\
\textbf{Training time consumption}: We use one NVIDIA Titan RTX to count the average time consumption on Forward+Backward passes for one image for all CAR methods with $K=80$ as well as the original regression method and the Deep Ensembles~\cite{2017simple} using the same training settings.
\vspace{-1.5em}
\subsection{Performance and discussions}
\vspace{-0.5em}
Table~\ref{tab:fcn_depth_kitti} and Table~\ref{tab:bts_depth_kitti} provide depth and uncertainty results.\\
{\textbf{Depth: }}We find that all CAR MDE methods are portable, but training directly with the settings of the original backbones degrades performance. We discover that the Adabins, DS-SIDE and SORN~\cite{bhat2021adabins, ren2019deep,diaz2019soft} based models are more sensitive to the selected backbone than the other ones. Despite the influence of the backbones, we also consider that the training settings for the original Adabins are more different from the ones of BTS. This difference may cause Adabins to produce worse performance after porting.
% We report that the biggest difference is the batch size, where the original Adabins uses 4 times the batch size (16 images) than the original BTS (4 images).
DORN-based model~\cite{fu2018deep} achieves the best result among CAR DNNs, which confirms the effectiveness of ordinal constraints.\\
{\textbf{Uncertainty: }}Our proposed E-Dist shows good and robust performance in most cases given a CAR MDE method. Among the CAR strategies, we found that the uncertainty quality is related to the sharpness of the labeling during discretization, and also to the loss function. Li et al.~\cite{li2018monocular}, Yang et al.~\cite{yang2019inferring} and Cao et al.~\cite{cao2017estimating} based DNNs perform better for the uncertainty. Li et al.~\cite{li2018monocular} model has one-hot encoded labels in the classification map which leads to the sharpest label distribution. Yang et al.~\cite{yang2019inferring} model has $\gamma=15$ in Eq.~\ref{eq:softlabel1} and we can also have $\gamma \cdot q^{-2}=65$ in Eq.~\ref{eq:smooth3} for Cao et al.~\cite{cao2017estimating} model. 
This big coefficient can sharpen the label distribution. Conversely, in SORN~\cite{diaz2019soft} the $\gamma$ in Eq.~\ref{eq:softlabel2} is much smaller, which results in the evener distributed labels, and we consider this is the main cause of its worse performance. Yang et al.~\cite{yang2019inferring} based model outperforms the others, which indicates that the Multi-BCE loss is more suitable for uncertainty estimation, which is similar to the one-versus-all strategy~\cite{franchi21access}.\\
{\textbf{Choices of $K$: }}According to two sets of results separated by $K$, the performance rankings are consistent: the depth precision and the uncertainty quality increase along $K$, and $K$ has biggest impact on Cao et al.~\cite{cao2017estimating} based model.\\
{\textbf{Time efficiency: }}According to Table.~\ref{tab:time}, we argue that the CAR strategy will slightly slow down the training, especially for the
% the Li et al.~\cite{li2018monocular}, DS-SIDE~\cite{ren2019deep} and Adabins~\cite{bhat2021adabins} based models require relatively short training time, because the former two methods directly use one-hot to construct the target classification map in discretization, and the latter one does not need to make a specific target classification map. 
ones requiring label smoothing in discretization~\cite{yang2019inferring, cao2017estimating, fu2018deep}. 
However, Deep Ensembles~\cite{2017simple} with only three models still require the most training time. 
% While this can improve the accuracy of depth predictions, it does not provide the same quality of uncertainty as most CAR methods as we can see in Table.~\ref{tab:fcn_depth_kitti}.

\begin{table}
\renewcommand{\captionfont}{\footnotesize} 
\centering
\scalebox{0.45}{
\begin{tabular}{llllllllll}
\toprule
\multicolumn{10}{c}{\textbf{\textbf{Time consumption (ms)}}} \\ 
\toprule
\textbf{\textbf{\begin{tabular}[c]{@{}l@{}}CAR\\Solutions\end{tabular}}} & DORN~\cite{fu2018deep} &\begin{tabular}[c]{@{}l@{}}Cao\\et al.~\cite{cao2017estimating}\end{tabular} & Li et al.~\cite{li2018monocular} & SORN~\cite{diaz2019soft} & \begin{tabular}[c]{@{}l@{}}Yang\\et al.~\cite{yang2019inferring}\end{tabular} & DS-SIDE~\cite{ren2019deep} & Adabins~\cite{bhat2021adabins} & Org & \begin{tabular}[c]{@{}l@{}}Deep\\Ensembles~\cite{2017simple}\end{tabular} \\
\midrule
\textbf{\textbf{BTS}} & 610.96 & 509.18 & 431.32 & 444.66 & 613.98 & 430.14 & 421.06 & 378.98 & 1136.94 \\
\textbf{\textbf{FCN}} & 735.38 & 614.52 & 538.92 & 556.84 & 722.80 & 540.70 & 588.26 & 517.66 & 1552.98 \\
\bottomrule
\end{tabular}
}
\vspace{-0.5em}
\caption{Time consumption on Forward+Backward passes for one image.}
\label{tab:time}
\end{table}

%% file: 5supp.tex
\begin{center}
{\textbf{\large{On Monocular Depth Estimation and Uncertainty Quantification using Classification Approaches for Regression\\}}}
\textbf{\large{------ Supplementary Material ------}}
\end{center}

\begin{figure*}[h]
    %  \centering
     \begin{subfigure}[b]{0.24\textwidth}
         \centering
         \includegraphics[width=\textwidth]{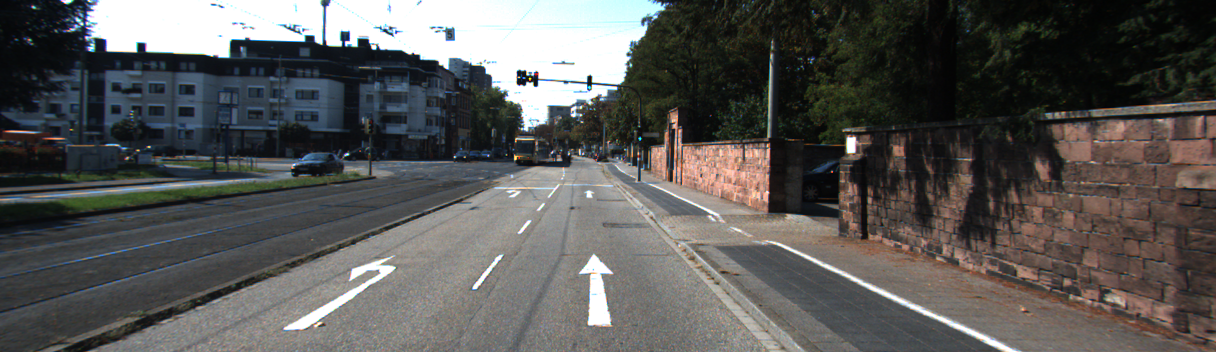}
         \caption{Input image}
     \end{subfigure}
    %  \hfill
     \begin{subfigure}[b]{0.24\textwidth}
         \centering
         \includegraphics[width=\textwidth]{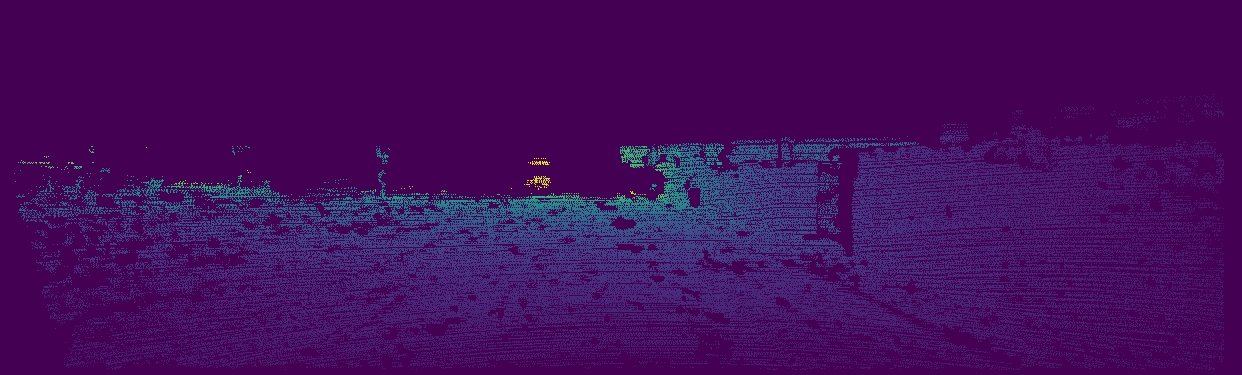}
         \caption{Ground truth depth}
     \end{subfigure}
    %  \hfill
     \begin{subfigure}[b]{0.24\textwidth}
         \centering
         \includegraphics[width=\textwidth]{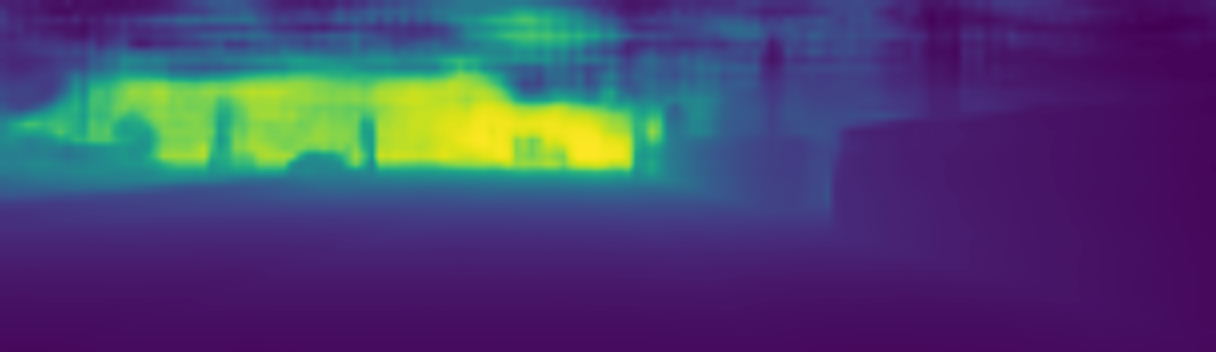}
         \caption{Predicted depth}
     \end{subfigure}
    %  \hfill
     \begin{subfigure}[b]{0.24\textwidth}
         \centering
         \includegraphics[width=\textwidth]{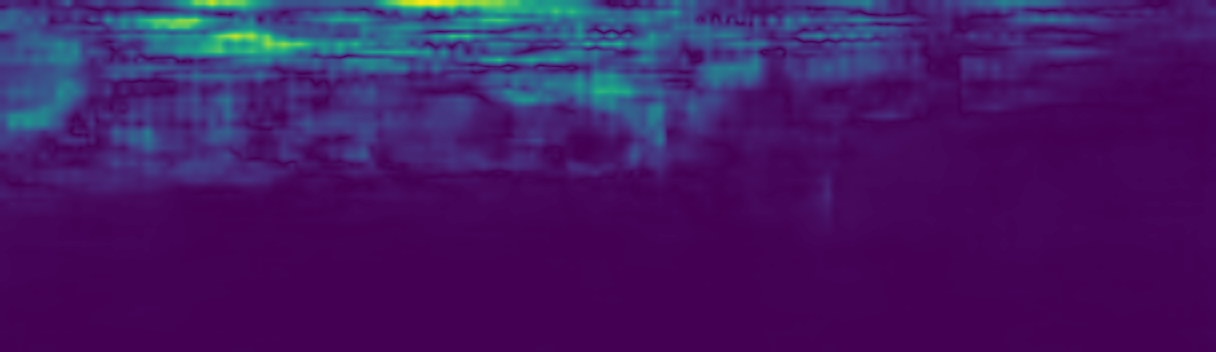}
         \caption{Variance uncertainty}
     \end{subfigure}

    \begin{subfigure}[b]{0.24\textwidth}
         \centering
         \includegraphics[width=\textwidth]{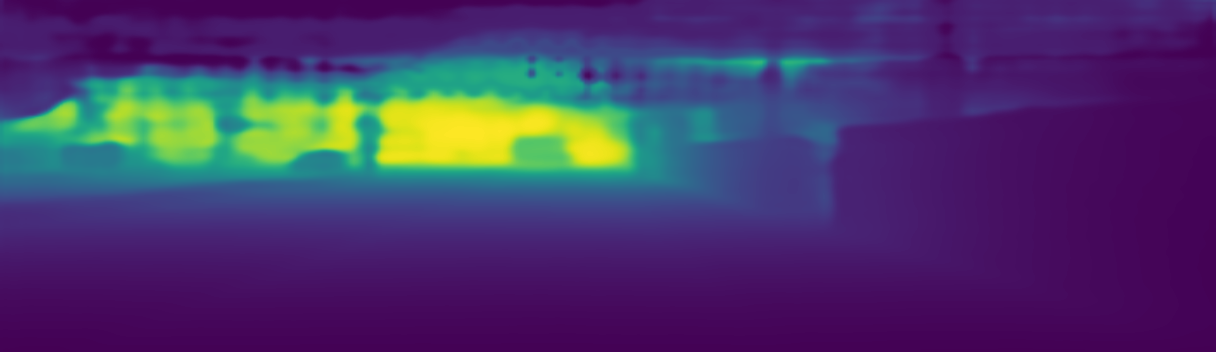}
         \caption{Predicted depth}
     \end{subfigure}
     \begin{subfigure}[b]{0.24\textwidth}
         \centering
         \includegraphics[width=\textwidth]{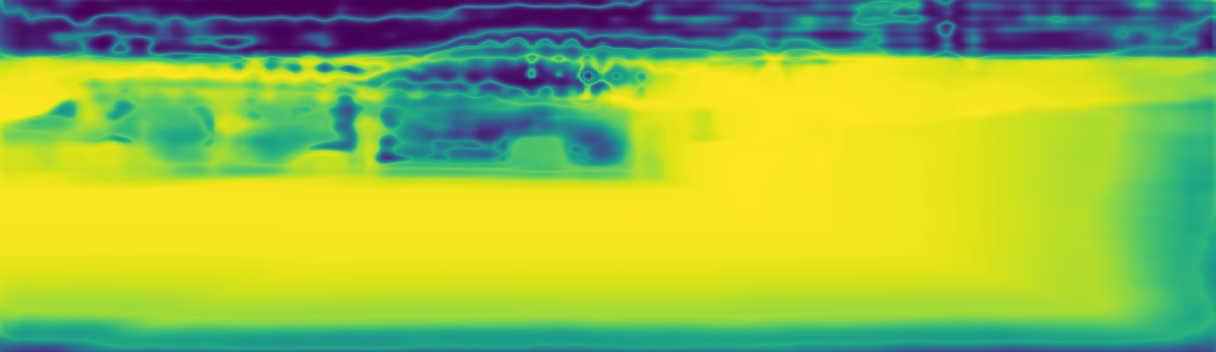}
         \caption{1-MCP uncertainty}
     \end{subfigure}
     \begin{subfigure}[b]{0.24\textwidth}
         \centering
         \includegraphics[width=\textwidth]{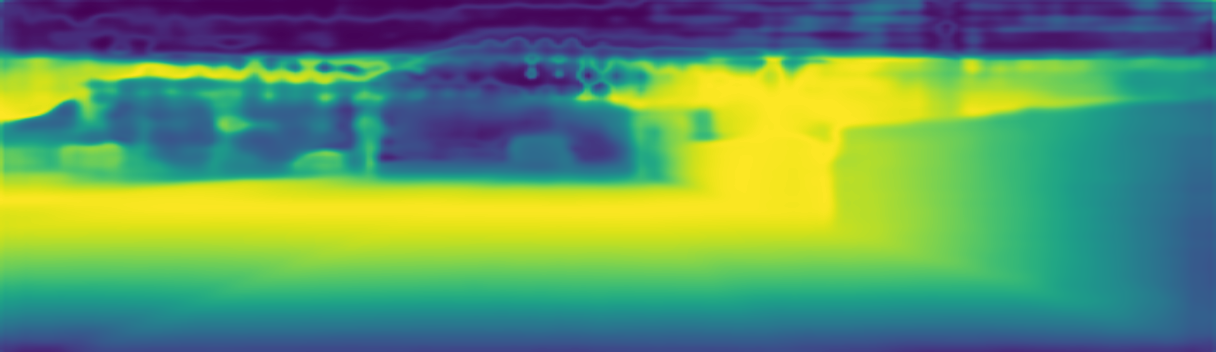}
         \caption{S-Entr uncertainty}
     \end{subfigure}
     \begin{subfigure}[b]{0.24\textwidth}
         \centering
         \includegraphics[width=\textwidth]{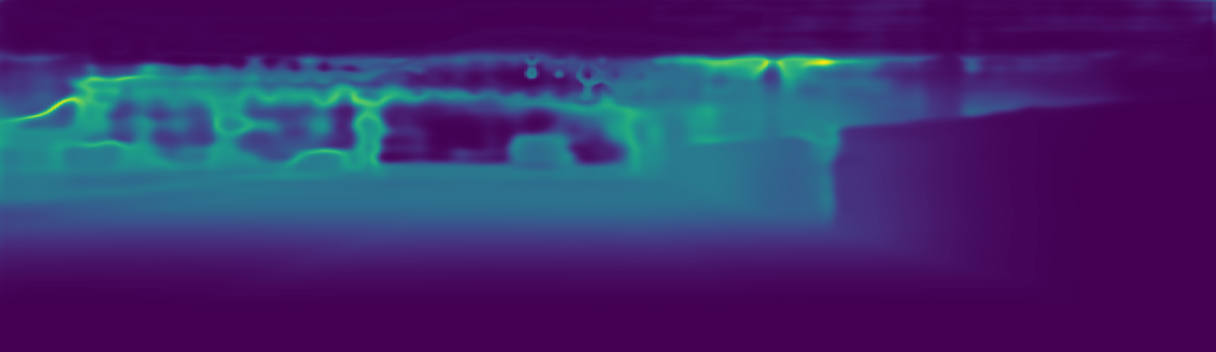}
         \caption{E-Dist uncertainty}
     \end{subfigure}

    \begin{subfigure}[b]{0.24\textwidth}
         \centering
         \includegraphics[width=\textwidth]{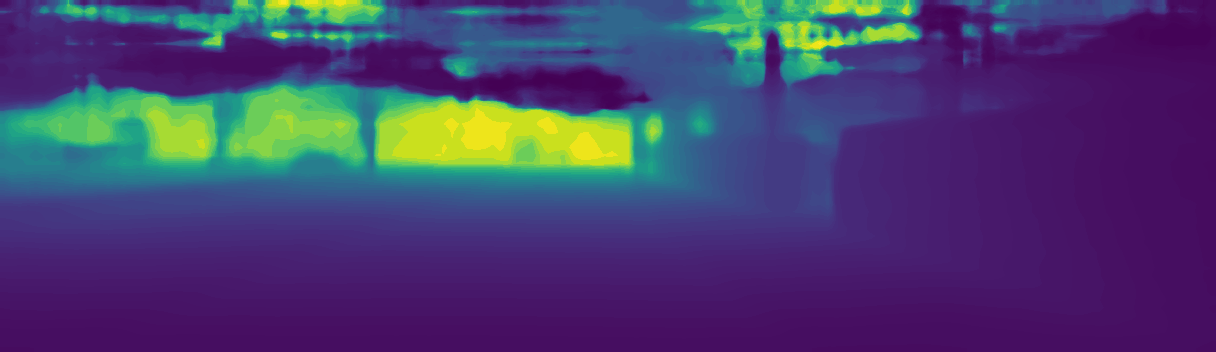}
         \caption{Predicted depth}
     \end{subfigure}
     \begin{subfigure}[b]{0.24\textwidth}
         \centering
         \includegraphics[width=\textwidth]{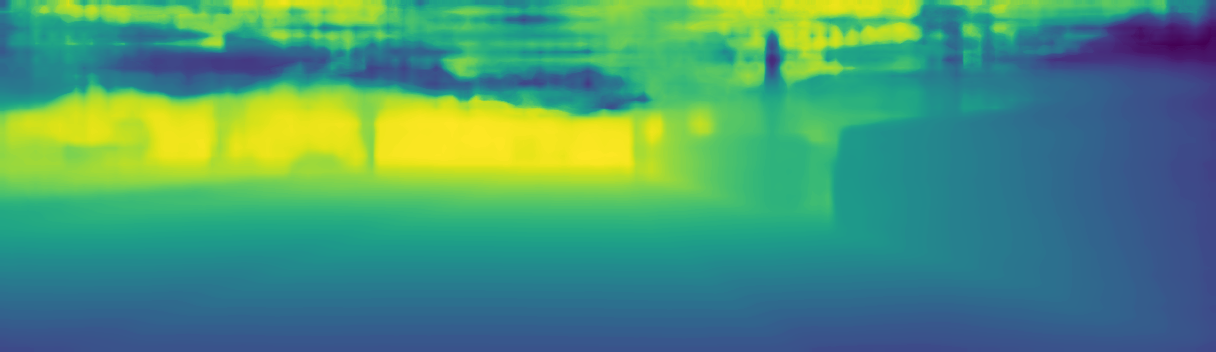}
         \caption{1-MCP uncertainty}
     \end{subfigure}
     \begin{subfigure}[b]{0.24\textwidth}
         \centering
         \includegraphics[width=\textwidth]{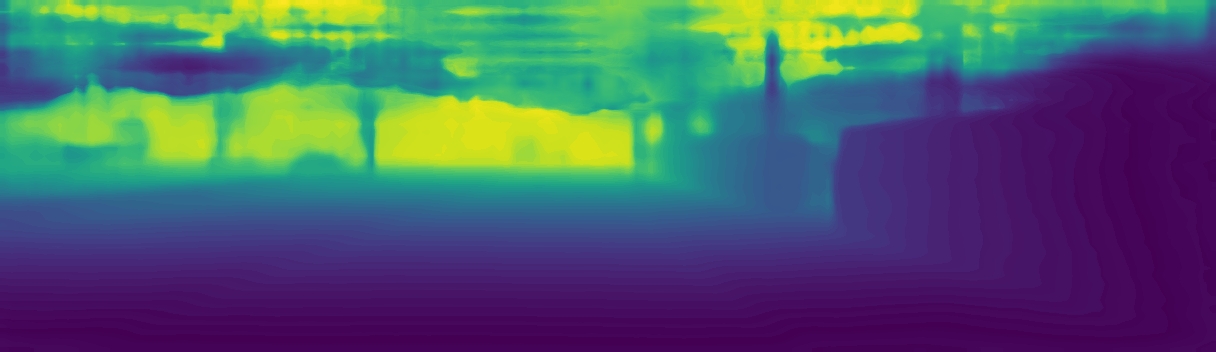}
         \caption{S-Entr uncertainty}
     \end{subfigure}
     \begin{subfigure}[b]{0.24\textwidth}
         \centering
         \includegraphics[width=\textwidth]{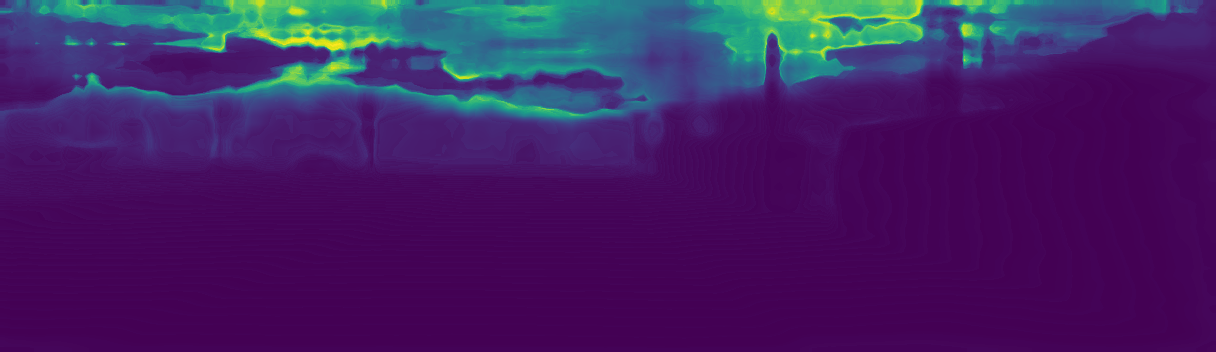}
         \caption{E-Dist uncertainty}
     \end{subfigure}

    \begin{subfigure}[b]{0.24\textwidth}
         \centering
         \includegraphics[width=\textwidth]{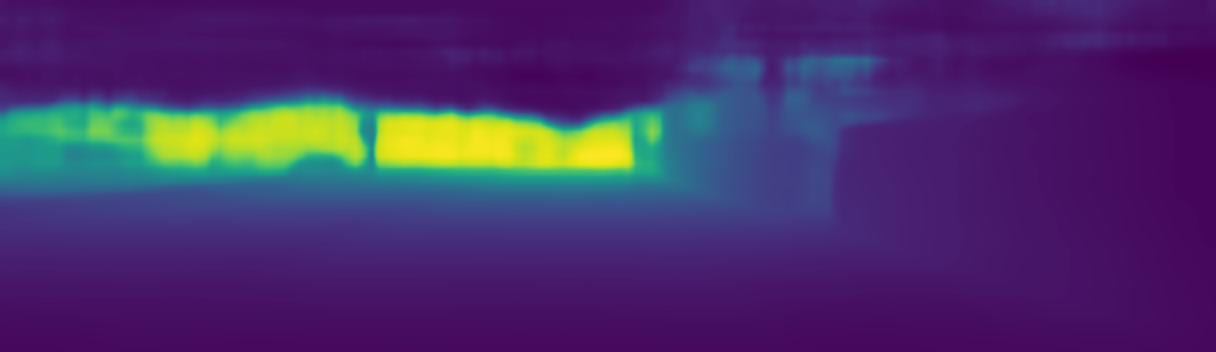}
         \caption{Predicted depth}
     \end{subfigure}
     \begin{subfigure}[b]{0.24\textwidth}
         \centering
         \includegraphics[width=\textwidth]{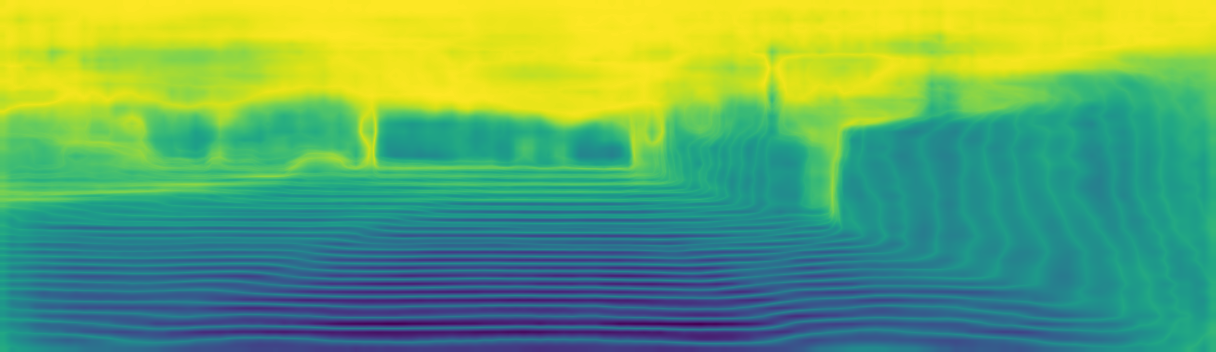}
         \caption{1-MCP uncertainty}
     \end{subfigure}
     \begin{subfigure}[b]{0.24\textwidth}
         \centering
         \includegraphics[width=\textwidth]{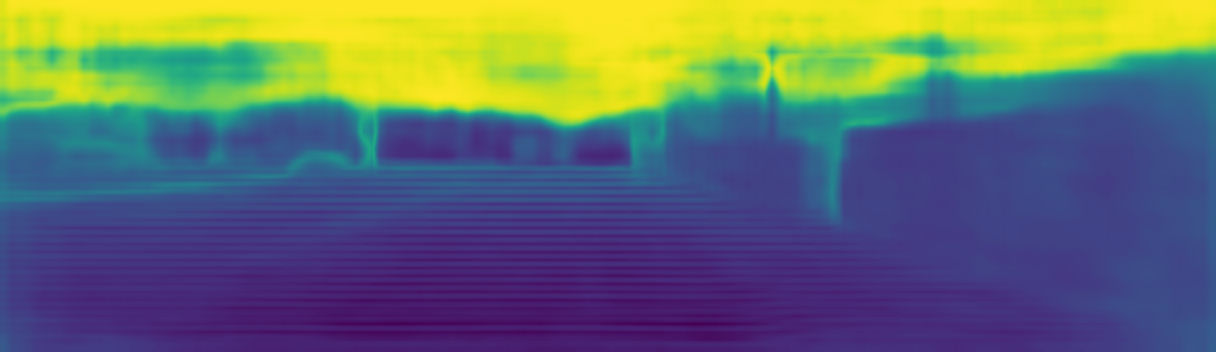}
         \caption{S-Entr uncertainty}
     \end{subfigure}
     \begin{subfigure}[b]{0.24\textwidth}
         \centering
         \includegraphics[width=\textwidth]{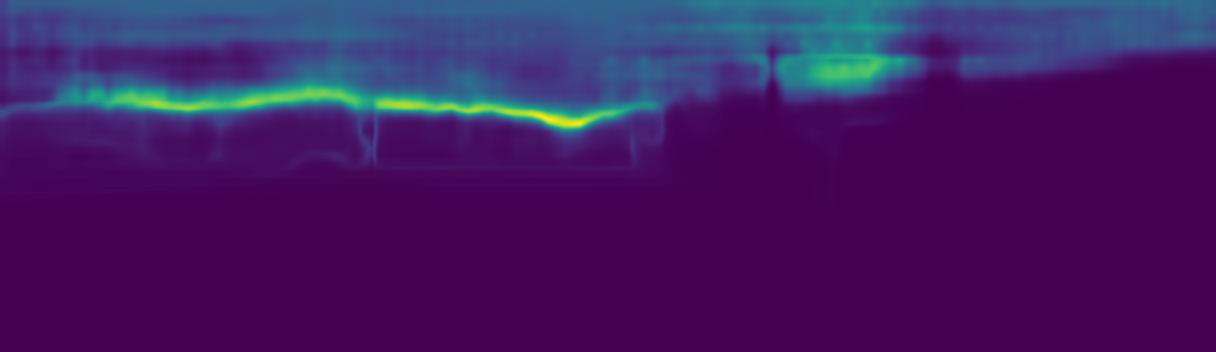}
         \caption{E-Dist uncertainty}
     \end{subfigure}
     
        \caption{Illustrations of predicted depth and uncertainty for the selected strategies applied on KITTI dataset. For both depth and the uncertainty, the brighter the pixel is, the higher depth/uncertainty value is. The figures are arranged as follows:\\\textbf{Input image and ground truth depth}: (a) (b). For the different strategies: \textbf{Deep Ensembles}~\cite{2017simple}: (c) (d); \textbf{Adabins based}~\cite{bhat2021adabins}: (e) - (h); \textbf{Dorn based}~\cite{fu2018deep}: (i) - (l); \textbf{Yang et al. based}~\cite{yang2019inferring}: (m) - (p). All theses strategies are based on FCN-ResNet101~\cite{long2015fully, he2016deep} backbone. The different uncertainty outputs are given by the solutions introduced in Sec. 2.2 in the main paper.}
        \label{fig:three graphs}
\end{figure*}

\section*{Overview}
\subsection*{Notations}
Table.~\ref{tab:notations} lists some of the notations we denote and use in the Sec. 2 in the main paper.
\begin{table}[h]
\centering
\scalebox{0.65}{
\begin{tabular}{lll} 
\toprule
\textbf{Types} &
\textbf{Notations} & \textbf{Meanings}\\
\toprule
\multirow{4}{*}{{\begin{tabular}[c]{@{}c@{}}sub/super-\\scripts\end{tabular}}} & $i$ & subscript for image/depth index\\
 &$j$ & subscript for pixel index\\
 &$p$ & subscript for channel index\\
 &${[\text{method name}]}$ & superscript for indicating different methods\\
\midrule
\multirow{2}{*}{{\begin{tabular}[c]{@{}c@{}}common\\capital letters\end{tabular}}} & $N$ & number of pixels with valid ground truth \\
 &$K$ & number of bins, the level of discretization \\
\midrule
\multirow{6}{*}{{\begin{tabular}[c]{@{}c@{}}some other\\notations\end{tabular}}} & $\mathbf{d}_i = \{d_{i,j}\}_j^N$ & ground truth depth values\\
 &$\Bar{\mathbf{d}} = \{\Bar{d}_p\}_p^K$ & handcrafted logarithm depth table\\
 &$\hat{\Bar{\mathbf{d}}}_i = \{\hat{\Bar{d}}_p\}_p^K$ & learned (adaptive) depth table for a given image\\
 &$q$ & width between two side-by-side bins in the depth table \\
 &$f_{\theta_1}(\mathbf{x}_i)$ & output logits of the DNN $f_{\theta_1}$\\
 &$\hat{{\mathbf{y}}}_i = \{\{\hat{{y}}_{i,j,p}\}_p^K\}_j^N$ & softmax output of the DNN $f_{\theta_1}$\\
\bottomrule
\end{tabular}
}
\caption{Reminder for the notations}
\label{tab:notations}
\end{table}

\subsection*{Original settings for CAR strategies on KITTI experiment}\label{sec:carsupp}
Table.~\ref{tab:exp_settings} lists the original model backbone choices and experiment settings for different CAR strategies on KITTI experiment.
For the coefficient used for label smoothing in Cao et al.~\cite{cao2017estimating}, according to our discussions in the end of Sec. 2.1.2, we can transfer the original coefficient (0.5) to 65 as we report in the table using $\gamma \cdot q^{-2} = 0.5 * (\log 80/50)^{-2} = 65$.

As we can see, the previous works collected in Table. 1 in the main paper lack a full comparison, and the network structures and hyperparameters they use are also different.

\subsection*{Complementary to the loss function descriptions}
\textbf{Multiple BCE loss}: is another solution when the target is a soft discrete distribution. Yang et al.~\cite{yang2019inferring} apply BCE loss on every class value in $\mathbf{y}^{[\text{smo1}]}_{i,j}$ defined in Eq.~\ref{eq:softlabel1}. The loss function is similar to Eq.~\ref{eq:ordinalloss}:
\begin{footnotesize}
\begin{align}
&{L_{i,j}^{{[\text{MBCE}]}}}(\theta_1) \text{=} -[ {{\mathbf{y}^{[\text{smo1}]}_{i,j}} \log \delta([f_{\theta_1}(\mathbf{x}_i)]_j) \text{+} (1 \text{-} {\mathbf{y}^{[\text{smo1}]}_{i,j}}) \log (1 \text{-} \delta([f_{\theta_1}(\mathbf{x}_i)]_j)} ]
\end{align}
\end{footnotesize}

\textbf{Regression loss (Smooth L1 loss)}: DS-SIDENet~\cite{ren2019deep} applies CAR with a smooth L1 loss~\cite{chang2018pyramid} to fit the one-hot classification map target:
\begin{align}
\centering
    &{L^{[\text{smoL1}]}_{i,j}}(\theta_1) =
    \left\{
        \begin{array}{rl}
            0.5(\hat{d}_{i,j} - k)^2  & {\text{if } |\hat{d}_{i,j} - k|<1}\\
            |\hat{d}_{i,j} - k| - 0.5  & {\text{otherwise}}
        \end{array}
    \right.\\
    &\text{with }\hat{d}_{i,j} = \sum_{p=0}^{K-1} \hat{{y}}_{i,j,p}\cdot (p+1)\nonumber
\end{align}
where $k$ is defined in Eq.~\ref{eq:onehot}. 

\textbf{Regression loss (Scale-Invariant loss)}: is applied along with a post-processing to produce the continuous depth. Adabins~\cite{bhat2021adabins} uses the per-image adaptive depth table $\hat{\Bar{\mathbf{d}}}^{[\text{ada}]}_i$ defined in Eq.~\ref{eq:learned2} instead of the fixed $\Bar{\mathbf{d}}$, then applies a Scale-Invariant loss~\cite{eigen2014depth}:
\begin{align}
\centering
    &{L^{[\text{SI}]}_{i}}(\theta_1,\theta_2) = \omega\sqrt{\frac{1}{N}\sum_{j=0}^{N}{{h^2_{i,j}}}-\frac{\lambda}{N^2}(\sum_{j=0}^{N}{h_{i,j}})^2}\\
    &\text{with }h_{i,j} = \log \hat{d}^{[\text{ada}]}_{i,j} - \log {d}_{i,j};\hspace{0.5em} \hat{d}^{[\text{ada}]}_{i,j} = \sum_{p=0}^{K-1} \hat{{y}}^{[\text{ada}]}_{i,j,p}\cdot {\hat{\Bar{{d}}}^{[\text{ada}]}_{i,p}}\nonumber
\end{align}
where $\omega$ and $\lambda$ are hyper-parameters.

\begin{table}[h]
\centering
\scalebox{0.6}{
\begin{tabular}{lcccccccc} 
\toprule
\begin{tabular}[c]{@{}c@{}}\textbf{CAR MDE}\\\textbf{solutions}\end{tabular}
 & \begin{tabular}[c]{@{}c@{}}\textbf{Encoder}\\\textbf{Backbone}\end{tabular}
 & \begin{tabular}[c]{@{}c@{}}\textbf{Choice}\\\textbf{of $K$}\end{tabular} & \begin{tabular}[c]{@{}c@{}}\textbf{Param.}\\\textbf{smooth}\end{tabular} & \begin{tabular}[c]{@{}c@{}}\textbf{Uncertainty}\\\textbf{eval.?}\end{tabular} & \begin{tabular}[c]{@{}c@{}}\textbf{Comp. with}\\\textbf{other CAR?}\end{tabular} & \begin{tabular}[c]{@{}c@{}}\textbf{Comp. with}\\\textbf{Reg?}\end{tabular} \\ 
\midrule
DORN~\cite{fu2018deep}&ResNet-101 & 80 & - & \XSolidBrush & \Checkmark & \Checkmark \\ 
\midrule
Cao et al.~\cite{cao2017estimating}&ResNet-101 & 50 &  65 & \XSolidBrush & \XSolidBrush & \Checkmark \\ 
\midrule
Li et al.~\cite{li2018monocular}&ResNet-152 & 50 (150 \cite{li2018deep}) &  - & \XSolidBrush & \XSolidBrush & \Checkmark \\ 
\midrule
SORN~\cite{diaz2019soft}&Xception & 120  & 1 & \XSolidBrush & \XSolidBrush & \Checkmark \\ 
\midrule
Yang et al.~\cite{yang2019inferring}&ResNet-50 & 128   & 15 & \Checkmark & \Checkmark & \Checkmark \\ 
\midrule
DS-SIDE~\cite{ren2019deep}&Self-made & 80 & - & \XSolidBrush & \Checkmark & \Checkmark \\
\midrule
Adabins~\cite{bhat2021adabins}&EfficientNet & 256 & - & \XSolidBrush & \Checkmark & \Checkmark \\
\bottomrule
\end{tabular}
}
\caption{Summary on KITTI experiment settings for typical CAR strategies. \textbf{Param. smooth}: Coefficient used for label smoothing. \textbf{Uncertainty eval.}: Whether this work evaluates the uncertainty and compared with the other works. \textbf{Comp. with other CAR}: Whether this work compared their proposed CAR strategy with the other CAR methods. \textbf{Comp. with Reg.}: Whether this work compared the CAR with the regression version of its model.}
\label{tab:exp_settings}
\end{table}

\section*{Experiments}
\label{sec:experiments}
\input{experiments_supp}

%% file: experiments_supp.tex
\subsection*{{Evaluation matrices}}
For the depth evaluation, we use the same matrices first introduced in~\cite{eigen2014depth} and used in many subsequent works. We list them as follows:
\\1. {RMSE}: $\sqrt{\frac{1}{|D|}\sum_{\mathbf{d}_i\in{|D|}}||\hat{\mathbf{d}}_i - \mathbf{d}_i||^2}$; 2. {Absrel}: $\frac{1}{|D|}\sum_{\mathbf{d}_i\in{D}}|\hat{\mathbf{d}}_i - \mathbf{d}_i|/\mathbf{d}_i$; 3. {Threshold dk}: Inlier metrics, $k$ in $dk$ indicates the power of the threshold ($t$), we take $t = 1.25$ and $dk = \frac{|A|}{|D|}, \text{where } A = \left \{ \mathbf{x}_i, \text{such that } \delta_i = \max(\frac{\hat{\mathbf{d}}_i}{\mathbf{d}_i}, \frac{\mathbf{d}_i}{\hat{\mathbf{d}}_i}) \text{ and } \delta_i < t^k \right \}$; 4. {SqRel}: $\frac{1}{|D|}\sum_{d_i\in{D}}{||\hat{\mathbf{d}}_i - \mathbf{d}_i||^2/\mathbf{d}_i}$;\\5. {RMSElog}: $\sqrt{\frac{1}{|D|}\sum_{d_i\in{|D|}}{||\log\hat{\mathbf{d}}_i - \log \mathbf{d}_i||^2}}$; 6. {log10}: $\frac{1}{|D|}\sum_{\mathbf{d}_i\in{|D|}}{|\log_{10}\hat{\mathbf{d}}_i - \log_{10}\mathbf{d}_i|}$.

\subsection*{Illustrations}
Fig.~\ref{fig:three graphs} shows some illustrations for predicted depth as well as the predicted uncertainty given by different uncertainty estimation strategies.